\documentclass[sigchi,nonacm,table]{acmart}
\renewcommand\footnotetextcopyrightpermission[1]{} 
\acmConference{}{}{}
\usepackage{tikz}
\usepackage{svg}
\usepackage{pdfpages}
\usepackage{makecell}
\definecolor{Gray}{gray}{0.85}
\definecolor{White}{gray}{0.0}
\definecolor{open_color}{RGB}{0, 185, 55}
\definecolor{open_plus_color}{RGB}{255, 122, 105}
\definecolor{spam_color}{RGB}{85, 159, 255}
\usepackage{colortbl}

\definecolor{Gray}{gray}{0.85}

\newdimen\satlevel
\newdimen\satdiameter
\newcommand{\like}[2][]{%
    \satdiameter=1.25ex\relax
    \ifcase#2\relax
        \satlevel=0pt\relax
    \or
        \satlevel=0.125\satdiameter
    \or
        \satlevel=0.225\satdiameter
    \or
        \satlevel=0.32\satdiameter
    \or
        \satlevel=0.5\satdiameter
    \fi
    \tikz[baseline=-0.3\satdiameter]{%
        \draw[#1] (0,0) circle (0.5\satdiameter);
        \fill[#1] (0,0) circle (\satlevel);
    }%
}

\usepackage{rotating}
\usepackage{longtable}
\usepackage{url}
\usepackage{multirow}

\AtBeginDocument{%
  \providecommand\BibTeX{{%
    \normalfont B\kern-0.5em{\scshape i\kern-0.25em b}\kern-0.8em\TeX}}}

\usepackage[latin1]{inputenc}
\definecolor{light-gray}{rgb}{0.8,0.8,0.8}

\setcopyright{none}
\begin{document}

\title{A Neophyte With AutoML: Evaluating the Promises of Automatic Machine Learning Tools}

\author{Oleg Bezrukavnikov}
\email{obezrukavnikov@gmail.com}
\affiliation{
  \institution{Letovo School}
  \city{Moscow}
}
\author{Rhema Linder}
\email{rlinder@utk.edu}
\affiliation{
\institution{University of Tennessee}
  \city{Knoxville}
  \state{Tennessee}
}

\begin{abstract}

This paper discusses modern Auto Machine Learning (AutoML) tools from the perspective of a person with little prior experience in Machine Learning (ML). 
There are many AutoML tools both ready-to-use and under development, which are created to simplify and democratize usage of ML technologies in everyday life. 
Our position is that ML should be easy to use and available to a greater number of people.
Prior research has identified the need for intuitive AutoML tools.
This work seeks to understand how well AutoML tools have achieved that goal in practice.
We evaluate three AutoML Tools to evaluate the end-user experience and system performance.
We evaluate the tools by having them create models from a competition dataset on banking data.
We report on their performance and the details of our experience. 
This process provides a unique understanding of the state of the art of AutoML tools.
Finally, we use these experiences to inform a discussion on how future AutoML tools can improve the user experience for neophytes of Machine Learning.

\end{abstract}



\begin{teaserfigure}
\end{teaserfigure}

\maketitle

\section{Introduction}


While smaller datasets can be directly manipulated and analyzed by a real person in order to discover some insights \cite{zgraggen2016tableur}, AutoML tools are designed for very large datasets, which are very difficult to understand and interpret and often require indirect analysis.
As datasets become bigger in size and more complex in terms of data stored, Machine Learning techniques get more useful and even necessary for analysis. 
These techniques help to better interpret big data without the need of detailed analysis of all data points by transmitting most of the intense work to the computer. 
Some kinds of datasets require ML techniques.
For example, datasets with large number of images over the web \cite{deng2009imagenet,rerabek2016new}, information on behaviour of users of different apps \cite{granka2004eye}, and information on people's preferences \cite{agichtein2006learning}.


Ideally, a neophyte developer could create Machine Learning models fast and easily. 
However, creating Machine Learning models is difficult, often requiring a lot of math expertise and hours of intense labor.
As well, there are many datasets with a very large amount of data.
Although new techniques \cite{belabbas2009spectral} help ML-engineers to deal with such type of data faster, it still requires a lot of work.
To reduce the cost of ML analysis, prior research has developed a variety of AutoML tools.
These new tools are created to simplify and democratize \cite{shang2019democratizing} the usage of ML technologies in everyday life.


In this paper we discuss some of the modern AutoML tools from the perspective of a person with little prior experience, known as a ``neophyte'', in Machine Learning.
These tools attempt the lofty goal of giving users an experience that ``automatically composes a Machine Learning pipeline \cite{shang2019democratizing}.''
However, the level of automation depends on the User Experience of AutoML tools.
We recognise User Experience (UX) as ``the subjective relationship between user and application \cite{calvillo2015assessing,mccarthy2004technology}.''
Using specialized algorithms, AutoML tools compare different Machine Learning techniques while performing hyperparameter optimization \cite{feurer_hyperparameter_2019}.
This process of testing ranks technique-parameter combinations, giving the user a choice of well-performing Machine Learning models fitted to the particular dataset.


Our overall aim is to understand how AutoML tools perform in terms of User Experience and accuracy. We will discuss their performance, ease of use, and some other parameters through experimentation, which we ran on a number of datasets. As well, we provide our recommendations which are informed by our experiences using the tools.
This is an empirical approach.
The three AutoML tools, which we decided to test are TPOP, AutoKeras and AutoGluon. These tools are not the only ones in AutoML field, but we have chosen them as they are quite well known, affordable and high-level.
The present research evaluates the degree AutoML tools achieve their goal of automation and suitability for widespread use.
We contextualize AutoML related work, describe out methods for evaluation, show results, and finally end with a discussion.
Our discussion summarized our recommendations for researchers to address AutoML tools' functionality to best improve their user experience for neophytes of Machine Learning.

\section{Related Work}

Recent books \cite{AutoML} and surveys \cite{zollerbenchmark} of the whole of AutoML research provide a more complete understanding than we will address.
Our related work section highlights three major types of research.
Like our research, related papers also compare AutoML systems.
Prior research typically either introduces a new benchmark for AutoML systems comparison, introduces a new AutoML system, or shares results of an AutoML challenge.
While both our paper and the prior work compares numeric scores among AutoML systems, our research also focuses on UX.  

\textbf{Papers introducing new benchmarks.}
A large body of research work focuses on inventing benchmarks \cite{gijsbers2019open,balaji2018benchmarking}.
This approach helps compare different AutoML systems with quantified accuracy relative to state-of-the-art and baseline methods.
This research introduces a new benchmark, explains it, and also quantifies performance of selected AutoML models on this benchmark. 
In their paper, Gijsbers et al. share a benchmark for comparison of some of the most modern AutoML tools like TPOT \cite{olson2019tpot}, Auto-WEKA \cite{thornton2013auto}, auto-sklearn \cite{feurer2019auto}, and H20 AutoML \cite{cook2016practical}.
Their benchmarks are open source and designed to be updated as AutoML tools change. 
Gijsbers et al. have worked with AutoML communities to maintain interoperability and fairness.
Balaji and Allen focus on the benchmark methodology, explaining how their benchmark was created and proving its high quality.
They also refer to the OpenML datasets \cite{gijsbers2019open}, which are presented in the earlier paper, and discuss quality issues, such as missing target features.
These problems often make programmatic comparisons among AutoML tools difficult to impossible.
In this paper, while we are not introducing new benchmarks, we are also performing tests on data to compare User Experience. 

\textbf{Papers introducing new AutoML systems}.
Despite the fact that automated methods for generating models is a long-standing problem, dating from before 1990 (as for example \cite{NIPS1988_149}), it remains a popular and unsolved problem in many contexts.
There are many AutoML systems currently in use, as well those under development, that try to solve these long-standing problems.
For example, AutoGluon \cite{erickson2020autogluon} is an Amazon developed project.
It is used for Tabular Prediction, Image Classification, Object Detection, and Text Classification.
 
Northstar \cite{kraska2018northstar} is in active development, but unreleased for public use, by professors from MIT and Brown University. 
Kraska et al.'s Northstar is an AutoML tool that goes beyond creating models inspired by previous systems \cite{crotty2015vizdom}.
Northstar integrates visualization systems that show connections in data, helping analysts manipulate presentations to better understand their data.
This solution seeks to make Machine Learning more accessible to an average person.
We agree with Kraska et al. that UX and simplicity are very important in Machine Learning field. 
Therefore, in our paper, we focus on the UX of contemporary AutoML tools.

\textbf{AutoML challenges}.
Competitions and challenges are a popular activity for people in the Machine Learning community.
Popular examples include the following platforms: 
\begin{itemize}
    \item https://www.kaggle.com/competitions
    \item https://www.drivendata.org/competitions/
    \item https://ods.ai/competitions
    \item https://mlcontests.com/.
\end{itemize}

Many of the challenges are sponsored by companies such as Google, Amazon and etc.
Competitions often offer winners money or free software access as a reward.
The main point of these challenges is that they help to boost research and awareness in certain fields of Machine Learning \cite{goodfellow2013challenges}. 
This results in finding better solutions and developing new technologies for big companies, but not always smaller companies and individual programmers.

The concept of an AutoML challenge is less mature than established Machine Learning challenge venues.
One of the biggest AutoML challenges is ChaLearn \cite{guyon_analysis_2019}.
It is mostly popular for competitions in the field of image processing and analysis.
In their paper, Guyon et al.~analyses data from different ChaLearn challenges, identifying possible strategies, scoring metrics, and results.
Despite this effort, the participants creating AutoML systems for the challenge tended to pay little attention to User Experience.
This demonstrates that there is a lack of work comparing UX across AutoML tools.
Our paper address this deficiency by concentrating on the User Experience rather than system performance.

\section{Methods}


This methods section outlines how we selected AutoML tools, our dataset, metrics for performance comparison, and how we evaluated user experience.

\subsection{Tools Selection}
Our reasoning for evaluating AutoML tools was to better understand their potential for helping people who are not ML specialist to benefit from modern techniques.
We selected AutoML tools that were both popular and recently developed.
The three AutoML tools, which we decided to test are TPOT \cite{olson_tpot_2019}, AutoKeras \cite{jin2019auto} and AutoGluon \cite{erickson2020autogluon}. 
These tools are not the only ones in the AutoML field, but we have chosen them because they are well known, affordable, and functional. 
AutoGluon is the most recent tool we are testing.
However, there are many AutoML tools under development.
For example, an unreleased system developed by Kraska et al.~supports automated data processing, data visualisation, and ML models \cite{kraska2018northstar}.

\subsection{Selected Tools}
The first public version of TPOT was released in 2015. 
It is one of the earliest AutoML tools.
TPOT was developed at the Computational Genetics Laboratory of the University of Pennsylvania with the goal of maximizing classification accuracy on supervised tasks \cite{schrider2018supervised}.
It is based on the scikit-learn Machine Learning Python library \cite{hackeling2017mastering}.
TPOT \cite{koza1992genetic} uses a combination of genetic programming \cite{koza1992genetic} and scikit-learn to provide effective models to users, based on performance with their data. 
TPOT's primary features are solving Classification and Regression problems, but may also provide support for Neural Networks.

AutoKeras was published in 2017 by DATA Lab at Texas A\&M University \cite{jin2019auto}.
Jin et al.~released AutoKeras with novel search strategies that find neural network configurations with the help of Bayesian optimization.
Further, AutoKeras is designed to adapt to different hardware to optimize its performance.
It is based on a Deep Learning library called Keras.
When creating a solution model for user, AutoKeras provides different Neural Network approaches from the Keras library.
AutoKeras provides users with Classification and Regression tools for structured data as well as images and text.

Amazon released the public version of AutoGluon in 2019.
AutoGluon uses both Machine Learning and Deep Learning algorithms to generate an effective model for users.
Its main goal is to democratize ML by decreasing the amount of time and resources required for training.
Unlike most other AutoML frameworks that primarily focus on model hyperparameter selection, AutoGluon Tabular combines multiple models as ensembles.
Experiments ran by AutoGluon developers show that such an approach offers better use of allocated training time than the search of the best model.
While AutoKeras highlights support for Image and Text Regression, AutoGluon provides tools for Object detection that identify and locate targets embedded in larger pictures.

\subsection{Dataset Selection}
Competitions in ML can help early career data scientists learn to work with realistic data.
Our dataset is from a competition in which the lead author of this paper recently participated (\url{https://onti.ai-academy.ru/competition}).
We picked this because competition datasets are designed to present simulated real-world problems. 

For the competition we were given generated data, presenting expenses of bank clients through 3 years.
For each client expense transaction we were given the time and date when it was performed, the category of products bought and the sum of the transaction.
For the training data we were also given the age group of the clients.
The data placed each client in one of four age groups.
The task was to train a ML model to predict the age group of a client.
The score was counted as the number of correct answers divided by the number of questions.

We modified the dataset for the purposes of our study to limit computation time to be appropriate  for a consumer grade computer.
This dataset, after some basic operations on data, provided us with a dataset (30000 rows and 1015 columns).
We decided to cut the data to test the AutoML, because the approximate time of work on full data was too big.
It predicted us around 10h of work with all data, but we wanted to test under conditions of a single workstation.
We tested the tools on a dataset size of 5000x1015.
We expected the tools to complete model selection on this data in around an hour.
However, in practice the computation took more time than we expected. Therefore, we performed evaluation of tools on three training data sizes: 250, 500, 1,000 rows.

\subsection{Performance Evaluation}
In order to compare AutoML tools quantitatively, we established a procedure for testing performance on the selected data. This section describes operations we applied to the data before testing.

\subsubsection{Procedure}

In order to compare the performance of the tools, we used a consumer-grade computer with the following hardware:
\begin{itemize}
\item NVIDIA GeForce GTX 1050 Ti
\item Core i7, 7700HQ
\item RAM 24GB DDR4
\end{itemize}

All the tools were used under the same conditions, having only the Jupyter Notebook and command line running on the computer.
We close non-essential windows programs in order to prioritize computing power for training.

We performed the following for each test:
\begin{itemize}
\item Data Clearing
\item Features Addition
\item Model Creation
\item Train-test Splitting
\item Model Fitting
\item Score Evaluation
\end{itemize}

For \emph{Data Clearing}, we removed rows with NaN values. 
For \emph{Features Addition}, we added new features based on aggregate functions on existing numerical values. 
For \emph{Model Creation} we used TPOT and AutoGluon with default parameters.
We used AutoKeras with specified max\_trials parameter (max\_trials=5, max\_trials=20) as these greatly impact training time. 
This parameter defines how many Neural Network models from Keras library will be used for model creation on the data.
For \emph{Train-test Splitting}, we used the first N rows of data as training data and last M rows of data as test data.
This selection helped us to choose the same data for all the tools so that their performance evaluation is more fair.
Mirroring the programming competition, we included data  with  approximately equal portions of each class for training and testing.
For \emph{Model Fitting}, we used the fit function corresponding to each tool with its default parameters. This function starts the various model training processes, which are approached differently for different AutoML tools.
For \emph{Score Evaluation}, we used default evaluation functions that measure accuracy as the total correctly predicted target values over the total number of examples.

We limited our training data size to reflect the constraint of limited computational resources of a general developer with a consumer-grade computer.
We performed evaluation of tools on three training data sizes: 250, 500, 1,000 rows.
Each training data size had 1,015 columns.
We did not compare results for bigger number of rows due to time requirements.
The slowest tool required our computer to perform over two hours of training on 1000 rows, which we believe is a considerable time for an independent developer, who might have lower computational resources than we had.

\subsubsection{Metrics}
We compared the performance of different tools using the same metric as in the competition - accuracy (number of right guesses divided by number of answers). 
For example, a random guess would yield an accuracy of around 0.25 as there are 4 classes of similar size in the data.
Our best result during the competition was 0.6438.

\subsection{User Experience Evaluation}
We decided to compare AutoML tools by qualitatively analyzing aspects of the user experience they provide.

\subsubsection{Documentation.}
We sought to evaluate the documentation of the selected AutoML tools.
We found methods for evaluating documentation that focus on preserving continued open-sourced development \cite{lethbridge2003software, mccauley1996documentation}.
However, we did not find methods for evaluating documentation that focuses on early learners of Machine Learning.
Based on our own experiences, we decided to evaluate three characteristics of documentation, which we see as important features for AutoML tool users.
We developed these characteristics to focus on User Experience of end-users, rather than potential contributors.
They are:

\begin{itemize}
    \item \emph{Quick start tutorials} - help new users to start working with the library.
    \item \emph{Documented integrations} - integrations that help users to find new potentially helpful usecases of the library. 
    \item \emph{Detailed documentation} - helps advanced users utilize the library at its full potential.
\end{itemize}

The results section shows our analysis of each AutoML tool according to the characteristics we define above.

\subsubsection{Simplicity of the First Use}
We believe that the simplicity of the start of usage is a crucial part of user experience. 
Many tools and technologies do not receive their deserved attention because users find them hard to initially configure.
We compared the chosen AutoML tools by the differences in the first setup and number of lines of code required to start.
As ML models are often intended to be used in different projects, we also compared the simplicity of model export and import of the chosen AutoML tools as it is crucial for integration.

\subsubsection{Logs}
Logs help developers understand what programs are doing.
They are especially useful when anomalies occur \cite{fu2014developers}.
As ML principles of work are sometimes quite confusing to understand, good logs play a big role in making ML tools democratized\cite{shang2019democratizing} and more accessible.
We compared how good are logs of the AutoML tools at presenting information and how customizable is this information.

\section{Results}
We present numeric results from performance tests of the chosen AutoML tools and our analysis of their user experience.

\subsection{Performance Results}
In Table~\ref{tab:results1}, we present a comparison of performance of AutoGluon, AutoKeras, and TPOT on our data from ML competition.

Our results show AutoML tool have higher accuracy than random (0.25) chance.
TPOT is being quite accurate while taking more time to train, AutoKeras can be very fast with properly set max\_trials parameters (less the 2000 seconds to train on 1000 rows of data), and AutoGluon has the best results in accuracy and takes reasonable training time (0.597 accuracy trained on 1000 rows).
The baseline Catboost model outperforms all AutoML tools for 250 and 500 rows of training data in accuracy and requires a much less training time. 
Overall, AutoGluon has similar accuracy results to baseline.
AutoGluon outperforms the baseline model for 1000 rows of training data (accuracy 0.597 vs 0.586).
Our baseline model used parameters selected by professional data scientists.
AutoGluon performed with a high accuracy without the benefit of specialist's tuning.

\begin{table*}[h!]
\rowcolors{1}{white}{light-gray}
  \caption{Each row of this table lists an AutoML tool's accuracy and training time performance on a subset of the ML competition data in our study. Overall, only AutoGluon performed better than our manually-programmed baseline. }
  \label{tab:results1}
    \begin{tabular}{crrrr}
{ \bf Tool  }
&   { \bf Train Size (rows) }
&   { \bf Work time (seconds) } 
&   { \bf Test Size (rows) }
&   { \bf Accuracy }
\\
{ \bf Catboost baseline }
&   250
&   64
&   29000   
&   0.536
\\
{ \bf Catboost baseline }   
&   500
&   107
&   29000   
&   0.565   
\\
{ \bf Catboost baseline }   
&   1000
&   136
&   29000   
&   0.586
\\
{ \bf TPOT  }
&   250 
&   2038
&   29000
&   0.512
\\
{ \bf TPOT  }
&   500
&   5280
&   29000
&   0.550
\\
{ \bf TPOT  }
&   1000
&   9814
&   29000
&   0.568
\\
\makecell{\bf AutoKeras (max\_trials = 5)}
&   250 
&   375
&   29000
&   0.417
\\
\makecell{\bf AutoKeras (max\_trials = 20)}
&   500
&   1229
&   29000
&   0.450
\\
\makecell{\bf AutoKeras (max\_trials = 20)}
&   1000
&   1857
&   29000
&   0.513
\\
{ \bf AutoGluon }
&   250
&   169
&   29000
&   0.509   
\\
{ \bf AutoGluon }   
&   500
&   892
&   29000
&   0.552   
\\
{ \bf AutoGluon }
&   1000
&   2482
&   29000
&   0.597
\\
\end{tabular}
\end{table*}

\subsection{User Experience Results}

\subsubsection{Documentation}
In Table~\ref{tab:doctable}, we show the results of our analysis of documentation. 
It shows that AutoKeras pays very close attention to details, AutoGluon focuses mainly on quick start tutorials, and all the tools have different side services' integrations described in their documentation. 

\begin{table*}[h!]
\rowcolors{1}{white}{light-gray}
  \caption{This figure depicts AutoML tools' documentation analysis. Overall, TPOT has the least number of use cases and AutoKeras has the most detailed documentation.}
    \begin{tabular}{cp{4.5cm}p{4.5cm}p{4.5cm}}
{ \bf Tool  }
& { \bf Simplicity of start }
& { \bf Integrations featured } 
& { \bf Details in documentation }
\\
 { \bf TPOT } 
& Has a quick start tutorial for Classification and Regression on its GitHub page and in github.io documentation. Also features specific examples of usage.
& Has a tutorial on work with Pytorch and Dask on github.io.
& Has detailed documentation on Classification and Regression.
\\
 { \bf AutoGluon  }
& Has a quick start tutorial for all main usecases on the official website. Has tabular quick start tutorial on the GitHub page. 
& Presents guidilines for pytorch users and explains work with H2O algorithms
& Although it has documentation on some additional features, AutoGluon does not have detailed documentation on the main AutoML tools on its website. Has docs catalog on github, which might be used to generate documentation.
\\
 { \bf AutoKeras }
& Has a quick start tutorial for all main usecades on the official website. Has Image Classification quick start on the GitHub page.
& Explains usage of TensorFlow Cloud with autokeras. Has a tutorial on Trains integration.
& Has detailed documentation on all main classes and functions of the core features.
\\
\end{tabular}
  \label{tab:doctable}
\end{table*}

\subsubsection{Simplicity of the First Use}
One way to make tools more approachable for neophyte developers is to reduce the barrier to entry, making the first use as easy as possible.
When compared, all of the three AutoML tools had decent documentation and examples for first usage and required a small amount of code to repeat a basic example.
Yet, AutoGluon required the least amount of code, featuring different time saving functions.

Considering integration of models in different projects, AutoGluon and AutoKeras offered standard ML export options.
AutoGluon models could be exported and imported with the help of Joblib Python library "dump" and "load" functions or by using AutoGluon models' inbuilt "save" and "load" functions.
AutoKeras models could be exported and imported with a call of "save" and "load\_model" inbuilt functions.
TPOT did not present a simple export and import solution.
TPOT allowed users to save a model's training pipeline and use it in future model trainings, but there was no option to export and import the trained model itself.

\subsubsection{Logs}
\begin{figure*}[h!]
  \caption{This figure depicts example logging output from the three selected AutoML tools. One difference in logging is that AutoKeras reports on epochs, while TPOT displays evolving generations.}
  \centering
  \includegraphics[width=1\linewidth]{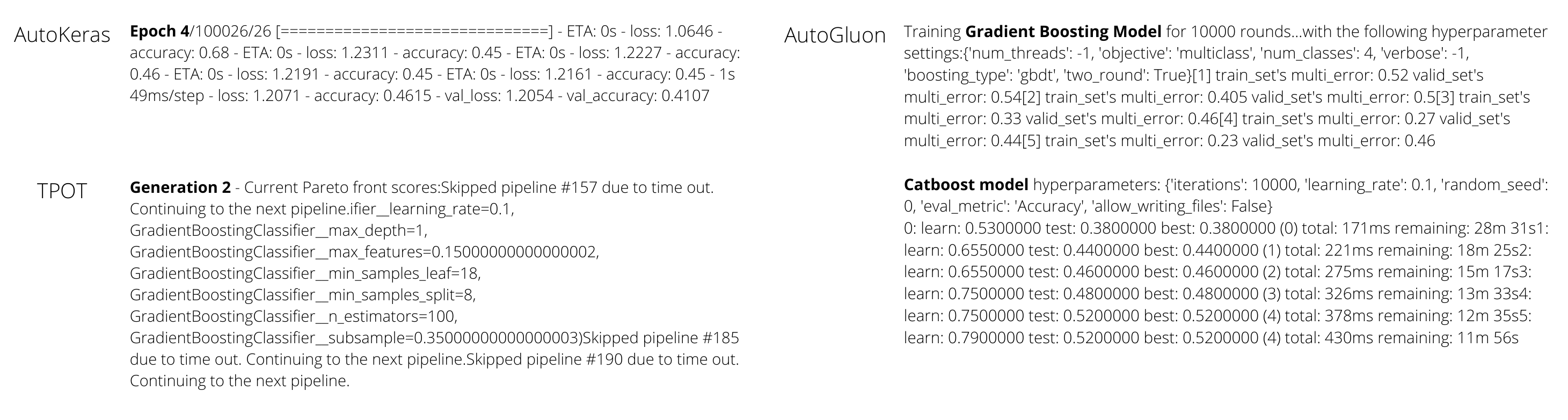}
  \Description{}
    \label{fig:logexamples}
\end{figure*}

When we used the selected tools, we found their logging to be quite different, see Figure \ref{fig:logexamples}.
TPOT and AutoGluon both incorporated a standard verbosity parameter \cite{li2017log}, which enables its users to choose an integer verbosity level that limits the level of details they see about ML training.
Popular programming tools like Hadoop often offer various levels of detail in their logging through \emph{verbosity parameters} \cite{li2017log}.
Some standard verbosity settings for software would be "ERROR", "WARNING", and "INFO".
These help system designers analyse program logic.

TPOT had 4 levels of verbosity and AutoGluon had 5.
In contrast, AutoKeras has some basic logs out of the box, but offers no verbosity parameter to customize the level of details.
Instead, it provides programming hooks where its users can make custom logs with tensorflow functions. 
While this feature might be particularly useful for advanced users, it is more demanding for newcomers. 
In Table~\ref{tab:logstable}, we describe the output of TPOT and AutoGluon, given the range of verbosity level settings.
At the lowest setting of verbosity, tools will not print out anything unless there are warnings.
At the highest verbosity setting, tools log detailed information on each training step.
The log information shows the current accuracy metrics and an overall progress indicator.

In Figure~\ref{fig:logexamples} we present examples of logs provided by AutoKeras, TPOT, and AutoGluon. 
While all of these tools have the same goal of providing a model, their logs are different because the nature of algorithms used are different for each tool.
As we mention in the \emph{ Tools Description} section, AutoKeras uses neural networks for predictions, TPOT uses genetic programming, and AutoGluon uses model stacking.
Correspondingly, the tools present log information on epochs, generations, and specific model training.
The \emph{ Epoch} is a part of neural network training during which the learning algorithm works through the entire training dataset one time and updates its parameters \cite{brownlee2018difference}.
In the logs of AutoKeras we can see how model prediction metrics vary from epoch to epoch.
The \emph{Generation} is a part of genetic programming, which includes a random mutation of the training ML model \cite{koza1992genetic}.
Each output about generation by TPOT provides us with information on data analysis and evolution progress.
AutoGluon performs training of different models on the given dataset.
In its logs we can see outputs of each specific model.
In this case, the models are the Gradient Boosting model and Catboost model.

\begin{table*}[h!]
\rowcolors{1}{white}{light-gray}
  \caption{This figure presents verbosity levels description for AutoGluon and TPOT. AutoKeras does not have standard logs with verbosity levels so we omit it from the table.}
    \begin{tabular}{rp{7cm}p{5cm}}
\\
{ \bf Verbosity level }
&   { \bf AutoGluon }
&   { \bf TPOT  }
\\
{ \bf 0 }   
&   Only prints exceptions.
&   TPOT prints nothing.
\\
{ \bf 1 }
&   Provides top level AutoGluon info (accuracy, directory, models).
&   Prints only warnings.
\\
{ \bf 2 }
&   Inclueds basic info about all models that are being trained.    
&   TPOT will print training results and provide a progress bar.    
\\
{ \bf 3 }
&   Prints information about info storage in addition to basic info.    
&   Includes detailed information about training and a progress bar.
\\
{ \bf 4 }
&   Presents time spent on various model trainings, their result, and accuracy. Details how models are trained and how data was interpreted. For example, it prints parameters used in classifiers, splits of train and test, and final scores.
&   No such option in TPOT.
\\
\end{tabular}
  \label{tab:logstable}
\end{table*}

\begin{figure*}[h!]
  \caption{This figure depicts example logging by AutoGluon. It shows both specific AutoGluon logs and logs of the model being trained by AutoGluon, in our case LightGBMClassifier.}
  \centering
  \includegraphics[width=1\linewidth]{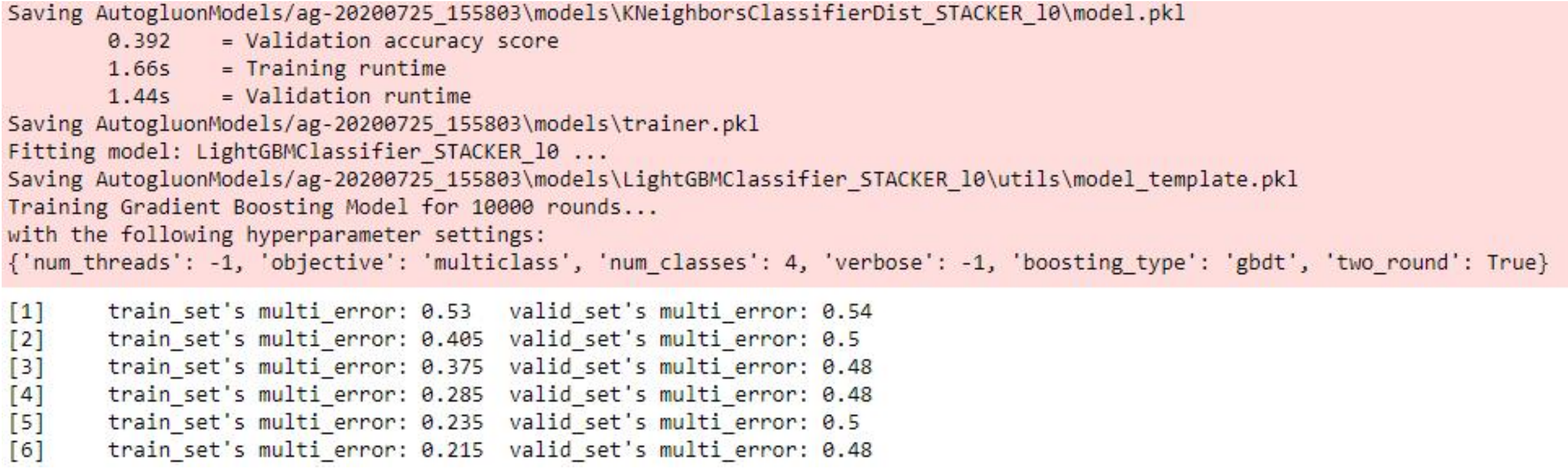}
  \Description{}
    \label{fig:autogluonlogs}
\end{figure*}

Overall, AutoGluon seemed to present the most useful logs.
In contrast, AutoKeras and TPOT logs were mixed up, confusing, and hard to interpret, AutoGluon keeps its logs more minimalistic.
It carefully utilizes line breaks, spaces, and colors to highlight the most important information, see Figure \ref{fig:autogluonlogs}.
In addition, AutoGluon provides clear summaries of each new model training after it is done.

AutoKeras and TPOT tend to print out all information without convenient line breaks.
These two tools also lack clear summaries of training, making it hard to understand how the model is developing.
TPOT also outputs hard to interpret information on how the data features are analysed and modified.
Because TPOT combines information across mutations, features, and iterations, the user can become easily overwhelmed.
The difficulty with showing too much log information is that it can hide more important parts, such as how much accuracy is increasing over time.
While TPOT shows all of its information to the user, the way it is presented without order or summaries makes it confusing.

\section{Discussion and Implications for Design}
In this section we present our assessment on the current state of AutoML. We also share some tips which we learned through work with different AutoML tools.

\subsection{AutoML Tools' Limitations}
This section outlines some flaws of AutoML tools we managed to discover.

\subsubsection{Error Handling}
As was mentioned in the Performance Evaluation section, once while using AutoKeras we received a Runtime error.
This error was not connected with our code, but only with AutoKeras algorithms.
Sadly, this type of error is commonly faced by many developers.
We believe that AutoML tools should make their inside algorithms error resistant.
This lack of visibility of the internal state is highly disruptive to the effort to democratize Machine Learning.
Instead, designers and researchers should strive to make tools more accessible for those without specific knowledge to interpret the errors and apply needed modifications to input data.
\subsubsection{Indecipherable Logs}
Out of all the tools we have evaluated, considering logs, AutoGluon seemed to provide the least confusing experience. AutoGluon displayed minimalistic output with carefully utilized line breaks and colors to highlight the most important information.
However, its approach to logs is not ideal.
We believe that logs of future AutoML tools should not only utilize colors and be very clean as AutoGluon's logs, but should also refer to model training.
Logs would be more beneficial if they emphasized how models develop over the course of training.

\begin{figure*}[h!]
  \caption{This figure presents an example of training process illustration. It was generated with TensorDash. The charts present loss and accuracy changes during the model training.}
  \centering
  \includegraphics[width=1\linewidth]{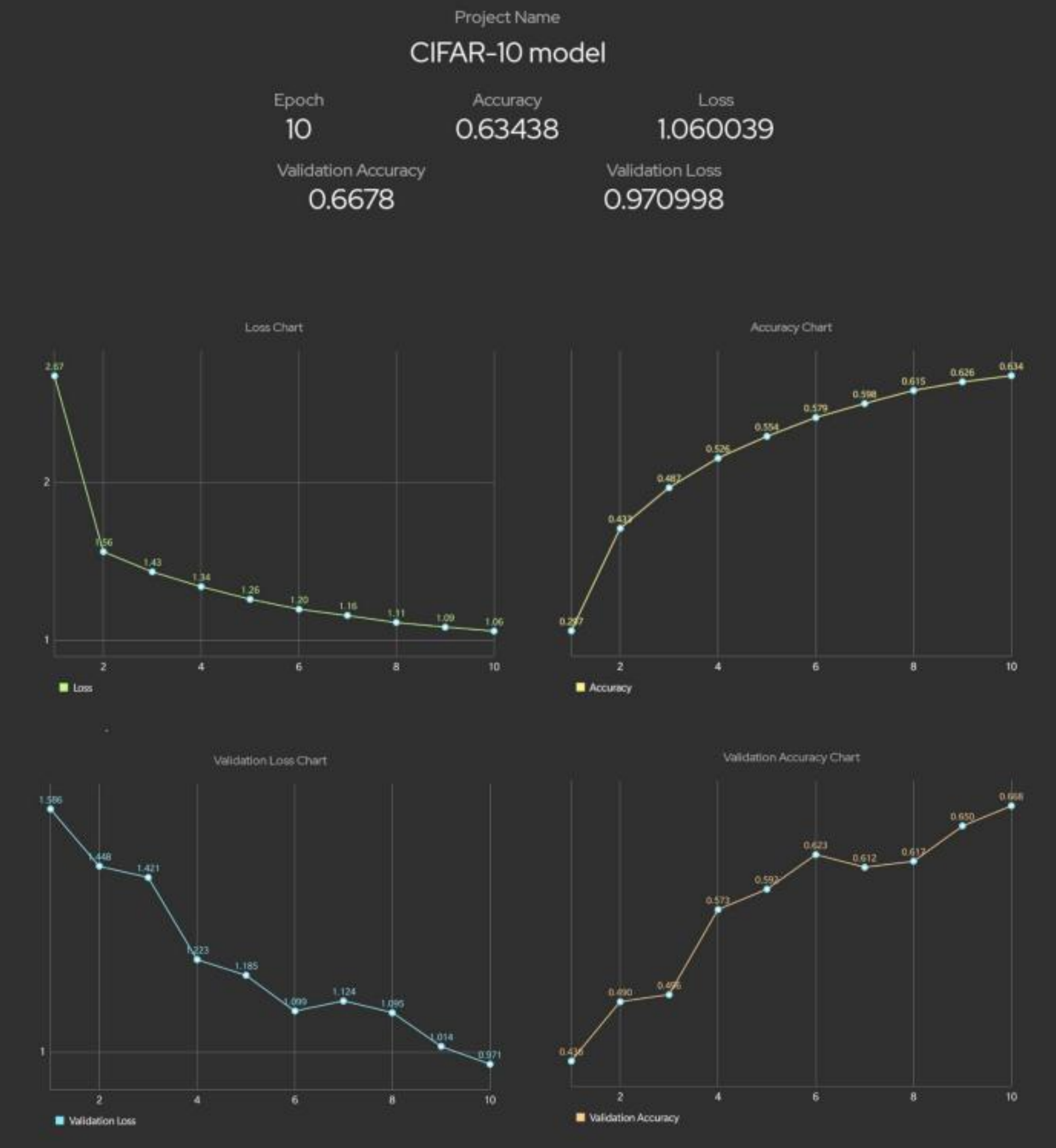}
  \Description{}
    \label{fig:visexample}
\end{figure*}

We highlight our experience that shows AutoML tools lack basic visualization support.
There are several Python visualisation libraries, which might be used to improve visual attractiveness and readability of logs including and not limited to matplotlib, plotly, pytables, ggplot \cite{Hunter:2007, stanvcin2019overview, fahad2018big}.
Several AutoML integrations for model training visualization are now available. Some of such integrations are mentioned in our section on \emph{Auxiliary AutoML Integrations}.
However, all external tools require additional time to learn and might require a usage fee. 
Out-of-the-box support that makes use of a Python visualisation library would make it easier to analyse model training and performance.
An example of possible helpful visualization is given in the Figure~\ref{fig:visexample}.
This visualization was created with Tensor Dash, an app and Python library that enables remote monitoring of Machine Learning model training developed by Harshit Maheshari and Arun Kumar.

\begin{figure*}[h!]
  \caption{This figure presents an example of training process illustration. It was generated with Weights\&Biases. The charts present loss and accuracy changes during the model training and number of epochs of current training trial for each training step.}
  \centering
  \includegraphics[width=1\linewidth]{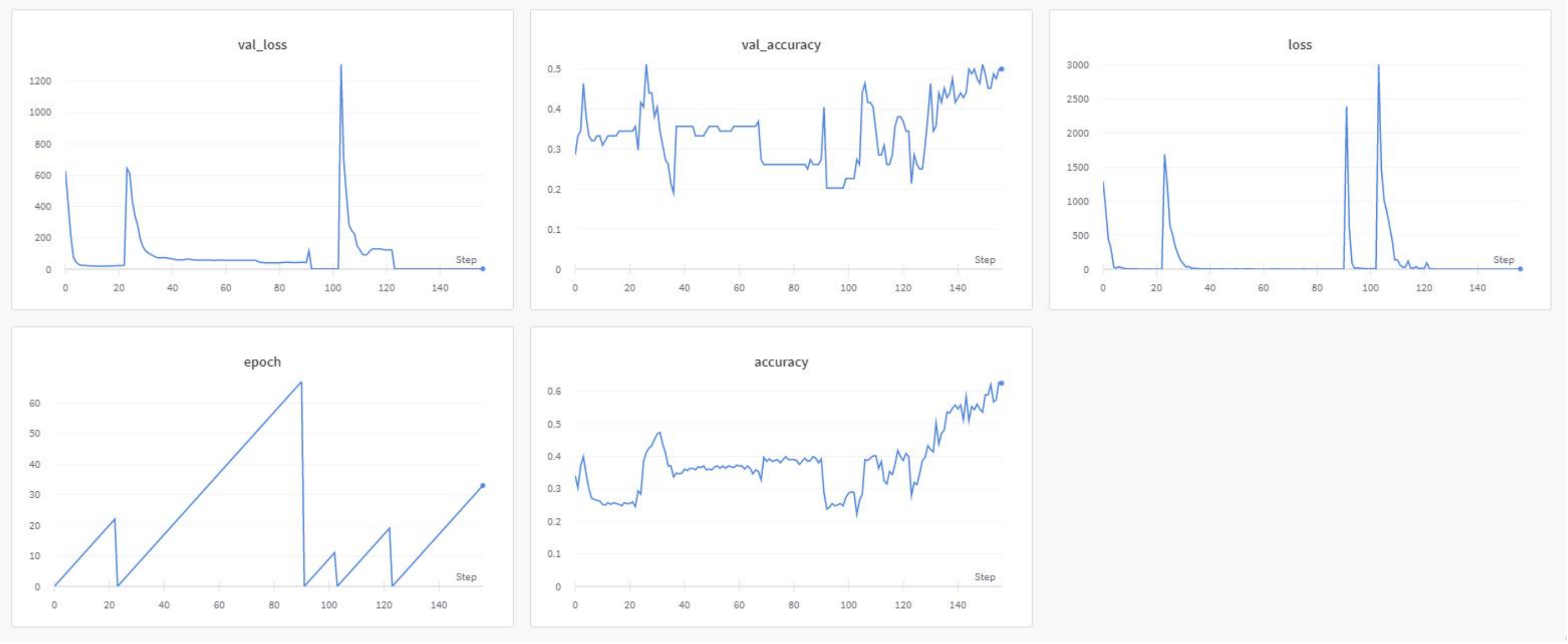}
  \Description{}
  \label{fig:wanb-dashboard}
\end{figure*}

\subsubsection{Time Estimation Problem}
One of the biggest barriers to entry to the ML field is the large amount of computational resources and machine time required to perform training on big data. 
Another source of friction into the ML field is the difficulty of knowing how many hours it will take to train.
For many companies and specialists it is necessary to know how long training will take.  
Training time differs based on computational resources, the nature of the data itself, the size of the data, and the tools chosen by the developer.
Better estimations on training time could help users schedule training and dedicate available resources at the most convenient time.
This would provide a better allocation of resources.
Some steps towards solution of this problem have been already offered by prior research \cite{justus2018predicting}.
However, there is still no readily available implementation.
Very large companies can absorb the cost of a lack of estimates because of their resources. 
But, this is not the case for small companies.
As more small companies and individuals enter the ML field, we predict that there will be a growing demand for training time prediction functionality.

\subsection{Building Towards Reduced Friction}
In this section we offer suggestions for creating a better user experience for future development of tools of AutoML tools.

\subsubsection{Auxiliary AutoML Integrations}
AutoML tools should continue to utilize auxiliary tools that simplify the user's needs.
There are several ML integrations tools which help tracking, automating, and analysing ML experiments.
Integrations include TensorDash, Weights\&Biases, Atlas, MLflow, TensorBoard, Comet, and Valohai.
These tools are designed to improve the experiences of developers during AutoML training.

In addition to the Performance Evaluation, we tested ML training visualization dashboards from Weights\&Biases and TensorDash.
TensorDash is a Python library and a mobile application, while Weights\&Biases is a Python library and a web-application.
TensorDash allows users to track their ML experiments from their phone.
Weights\&Biases provides users with an online dashboard for tracking, comparing, and analyzing their ML experiments.
We used both these tools together with AutoKeras with the help of the "callbacks" model's parameter.
An example of charts generated by TensorDash is given in the Figure~\ref{fig:visexample}.
One of the drawbacks of using TensorDash is that each TensorDash project contains charts for only one training.
As AutoKeras performs several trainings with different parameters, our projects' charts were constantly overwritten.
This left us without the results of training results across trials.
Weights\&Biases had a solution to this problem.
Weights\&Biases plots all data on the same chart, which allows you to examine all the trials together.
Some of the other advantages of Weights\&Biases over TensorDash are advanced chart scoping instruments and GPU usage charts.
An example of charts generated by Weights\&Biases is given in the Figure~\ref{fig:wanb-dashboard}.

Overall, we found TensorDash and Weights\&Biases easy to use. We especially see value in the speed with which its charts update during the training process.

\subsubsection{The Visual Future of AutoML}
The possible implementations of AutoML by developers is growing quickly.
While a skilled developer can produce good results, the tools are too complex for non-specialists.
Moreover, as the amount of data stored by different companies grows rapidly, the programmers and analysts are faced with challenges that grow in both size and complexity.
As Herbert Simon states \cite{simon1974big}, humans are best at operating on information precisely when it can fit in their cognitive memory.
However, human memory has limitations which are revealed as the amount of information for analysis increases.
When a Data Science related task involves an enormous amount of data, limitations of the human condition prevent analysts from keeping the data in their memory.
Therefore, systems which would offload their memory are necessary.

We highlight new efforts that utilize visual interfaces that intend to make AutoML more accessible for non-developers.
The Northstar project by Kraska et. al. \cite{kraska2018northstar} suggests a solution, which would enable anyone to create prediction models and find hidden dependencies between different data just by performing interactions with a simple user interface.
Built on the base of Vizdom, Northstar aims to provide a flexible interface that is friendly to both mouse and hand interaction.
This no code future of AutoML would highly democratize ML making its benefits accessible for everyone.
Our own experiences show that using AutoML tools is difficult, requiring expertise in Python, AutoML libraries knowledge, and expertise in tools' internal error handling.
Northstar's approach begins to remove these requirements.
The Northstar project intends to create a tool, which would allow general users to perform ML training with the help of simple table data and graph manipulation.
After model training, Northstar intends to provide users with generated Python code for prediction.
This code can be given to developers and applied in various pipelines.
By using the visual interface, Northstar fully manages code that allows users to perform various ML tasks without writing code directly.

As described by its creators, the main goal of the Northstar is "truly democratizing Data Science" \cite{shang2019democratizing}.
We agree with this goal.
The Northstar approach of reducing the amount of code though interactive applications is very promising.
While workers with various specializations lack programming skills, they are becoming more accustomed to using interactive applications.
As more tools that combine visual interfaces with ML backends are released, it will result in further democratization of ML development.
In the end, we believe this will make ML more ready for use by the general public.

\section{Conclusion}
We have analyzed three AutoML tools from the perspective of performance and user experience.
While autoML tools attempt ``automatically composes a Machine Learning pipeline \cite{shang2019democratizing}'', our goal was to explore whether this is indeed the case.
Instead of focusing on algorithm design and performance, we described our experiences utilizing AutoML tools features.
During this process, we encountered confusing log output, low-support for training time estimates, and a lack of support for visualizing internal state of training processes and results.
This work both details our experiences, documenting our experiences of Auto-ML state-of-the-art tools, and discusses recommendations for further improvement.
As democratization of ML continues, we hope that soon every interested developer will be enabled to take advantage of this amazing technology.

\bibliographystyle{ACM-Reference-Format}
\bibliography{sample-base}

\end{document}